\documentclass[twocolumn,10pt]{asme2ej}
\usepackage{amsmath}
\usepackage{epsfig} 
\usepackage{graphicx}
\graphicspath{ {./figure/} }
\usepackage{float}
\usepackage{graphicx}
\usepackage{caption}
\usepackage{subcaption}
\title{Nonlinear System Identification of Swarm of UAVs Using Deep Learning Methods}

\author{Saman Yazdannik
    \affiliation{
	MSc student\\
	Department of Aerospace Engineering\\
	K. N Toosi University of Technology\\
    Email: s.yazdannik@email.kntu.ac.ir
    }
}

\author{Morteza Tayefi
    \affiliation{
	Assistant Professor\\
	Department of Aerospace Engineering\\
	K. N Toosi University of Technology\\
    Email: tayefi@kntu.ac.ir
    }
}
\author{Mojtaba Farrokh
    \affiliation{
	Associate Professor\\
	Department of Aerospace Engineering\\
	K. N Toosi University of Technology\\
    Email: farrokh@kntu.ac.ir
    }
}

\begin{document}

\maketitle

\begin{abstract}
{\it This study designs and evaluates multiple nonlinear system identification techniques for modeling the UAV swarm system in planar space. learning methods such as RNNs, CNNs, and Neural ODE are explored and compared. The objective is to forecast future swarm trajectories by accurately approximating the nonlinear dynamics of the swarm model. The modeling process is performed using both transient and steady-state data from swarm simulations. Results show that the combination of Neural ODE with a well-trained model using transient data is robust for varying initial conditions and outperforms other learning methods in accurately predicting swarm stability.}
\end{abstract}


\section{Introduction}

The investigation and modeling of the interactions and dynamics of animal and robot swarms is a longstanding area of interest in academia. This area falls under the discipline of nonlinear System Identification (SID). Swarm dynamics are often nonlinear and governed by Ordinary Differential Equations (ODEs). Several data-driven SID methods exist, some of which will be discussed in the related work section.
Multiple swarm models exist and there is no unified model that governs all swarm systems. This study focuses on a specific type of swarm model from a particular class, with the general form described as.

\begin{align}
\dot{r_{i}}=v_{i}(t)                                    \\
\dot{v_{i}}=f(v_{i}(t)) - \sum_{j=1,i\neq j}^{N} g(r_{i}(t),r_{j}(t))+\eta_{i}(t)
\end{align}
\\
\\
\\

The swarm model under examination is governed by a second-order differential equation, where swarm behavior arises from acceleration ($v_{i}$) as outlined in Equation 2. It consists of three components: the intrinsic dynamics of an agent is dependent on its velocity, the interaction dynamics is a sum of functions of the agent's position and the positions of other agents, and model noise, typically assumed to be Gaussian white noise.

Models of this form can exhibit various steady-state behaviors including flocking, milling, and aggregation. Given time series position and velocity data obtained from observing a swarm at a fixed sampling rate and an unknown governing equation, the goal is to find an approximate function that accurately predicts the trajectories for any arbitrary initial conditions.

SID methods can be grouped into two categories. Given the dynamical process $\dot{x}=f(x(t))$ and the initial condition $x_{0}$, the first class of methods aims to directly predict future trajectory $x$, or to approximate $x_{0}+\int_{0}^{T}f(x(t))dt$, giving $x(T)$ for any T based on the initial state at $t = 0$. The second class of SID methods approximates $f(x(t))$ and then uses an ODE solver to integrate numerically and derive future $x(t)$ trajectories based on the initial state at $t = 0$. [RNN, CNN and MLP] belong to the first class while Neural ODE belongs to the second class, as evaluated in this study.

Due to inherent model noise, perfect prediction of all agents' future trajectories in a swarm is not feasible. The objective of this study is to identify models that accurately capture overall swarm behavior. A model is considered successful if it demonstrates correct convergence to the actual swarm's steady state and is robust to varying initial conditions. Experiments and metrics are presented to evaluate both criteria.

This study utilizes time-series data from simulations of a swarm model using a basic ODE solver based on Euler's method. The swarm's trajectory is divided into two phases: transient and steady state. With randomly initialized positions and velocities, a swarm initially enters the transient state before settling into the steady state. Results of training on data from both phases are presented. Time-series data prompted the consideration of MLPs, RNNs, and CNNs as initial models for this study.

\section{Modeling Swarm Dynamics in Nonlinear Systems}
We concentrate on a single swarm model from the previously mentioned swarm category.
\begin{align}
\dot{r_{i}}=v_{i}\\
\dot{v_{i}}=(1-|v_{i}|^{2})v_{i} -\frac{a}{N} \sum_{j=1,i\neq j}^{N} g(r_{i}(t),r_{j}(t-\tau ))+\eta_{i}(t)
\end{align}
In the present study, the intrinsic dynamics dictate that each agent experiences an acceleration or deceleration, leading to a constant velocity of 1 as time approaches infinity. The interaction dynamics cause each agent to accelerate towards the mean position of all other agents at a rate proportional to the coupling strength (a). It is assumed that Gaussian white noise is equivalent for all agents.

In this swarm model, three forms of stability are observed: "milling", "rotation", and "flocking". The first stability, "milling", is achieved under conditions of zero time lag, low noise, and an optimal coupling strength. The agents ultimately converge to a rough circular pattern around the mean field with a constant radius, resulting in a "milling" effect. This stability can manifest as all agents moving in a single direction or a mixture of clockwise and counter-clockwise movement. The second stability, "rotation", occurs when a significant time delay $(\tau)$ is introduced. The agents will eventually aggregate and rotate in a circle with a fixed radius. It is worth noting that simulations using Euler's method with large time steps are equivalent to introducing time delay to the swarm system. Our experiments show that, under specific model parameters, a swarm exhibiting "milling" stability can transition to "rotating" stability as the simulation time step is increased. The third stability, "flocking", emerges with the introduction of significant noise. The agents will all flock in a single direction. This study focuses on the learning of systems that exhibit the first two forms of stability.

The examination of model stability is a topic within the field of dynamical systems. This paper does not delve into the mathematical proof of model stability. Instead, we present empirical results obtained from our simulations. The second order ODE described in equation 4 can be reformulated as a first order ODE. As an example, consider a swarm consisting of 5 agents. The dynamics of each agent are governed by the second order ODE in equation 4. To simplify this ODE to a first order system, we introduce two variables for the $i^{th}$ agent: $x_{1i} = r_{i}$ and $x_{2i} = v_{i}$. This allows us to express the dynamics of the entire swarm as follows:
\begin{align}
    \dot{x_{11}}=x_{21},\\
    \dot{x_{21}}=(1-|x_{21}|^{2})x_{21} -\frac{a}{5} \sum_{j=1}^{5} (x_{11}(t)-x_{2j}(t))+\eta_{1}(t),\\
    \dot{x_{12}}=x_{22},\\
    \dot{x_{22}}=(1-|x_{22}|^{2})x_{22} -\frac{a}{5} \sum_{j=1}^{5} (x_{12}(t)-x_{2j}(t))+\eta_{2}(t),\\
    \dot{x_{13}}=x_{23},\\
    \dot{x_{23}}=(1-|x_{23}|^{2})x_{23} -\frac{a}{5} \sum_{j=1}^{5} (x_{13}(t)-x_{2j}(t))+\eta_{3}(t),\\
    \dot{x_{14}}=x_{24},\\
    \dot{x_{24}}=(1-|x_{24}|^{2})x_{24} -\frac{a}{5} \sum_{j=1}^{5} (x_{14}(t)-x_{2j}(t))+\eta_{4}(t),\\
    \dot{x_{15}}=x_{25},\\
    \dot{x_{25}}=(1-|x_{25}|^{2})x_{25} -\frac{a}{5} \sum_{j=1}^{5} (x_{15}(t)-x_{2j}(t))+\eta_{5}(t),
\end{align}

or we can write this as vector-matrix form:
\begin{equation}
   \begin{bmatrix}\dot{x_{11}} \\ \dot{x_{12}} \\ \dot{x_{13}} \\ \dot{x_{14}} \\ \dot{x_{15}} \\ \dot{x_{21}} \\ \dot{x_{22}} \\ \dot{x_{23}}\\ \dot{x_{24}}\\ \dot{x_{25}}\end{bmatrix} \\ = \begin{matrix} zeroes(5,5) & I(5,5) & zeroes(5,5)\\ \frac{a}{5}\cdot ones(5,5) -a \cdot I(5,5)   & I(5,5) & -I(5,5)\end{matrix} \begin{bmatrix}{x_{11}} \\ {x_{12}} \\ {x_{13}} \\ {x_{14}} \\ {x_{15}} \\ {x_{21}} \\ {x_{22}} \\ {x_{23}}\\ {x_{24}}\\ {x_{25}} \\ |x_{21}|^{2}{x_{21}} \\ |x_{22}|^{2}{x_{22}} \\ |x_{23}|^{2}{x_{23}} \\ |x_{24}|^{2}{x_{24}} \\ |x_{25}|^{2}{x_{25}}\end{bmatrix} + \begin{bmatrix}0 \\ 0 \\ 0 \\ 0 \\ 0 \\ 0 \\ 0 \\ 0\\ 0\\ 0 \\ \eta_{1} \\ \eta_{2} \\ \eta_{3} \\ \eta_{4} \\ \eta_{5}\end{bmatrix}  
\end{equation}

section{Literature Review on collective motion of a swarm}
\subsection{Literature Review on collective motion of a swarm}
The collective motion of a swarm can be understood as the interplay of three key components: short-range repulsion to prevent collisions, local interactions to align the velocity vectors of neighboring units, and a global positional constraint to maintain the coherence of the swarm [1]. A universal feature observed in collective swarm motion in biological systems such as schools of fish, bird flocks, and herds of mammals is the tendency for the velocity vectors of individual units to be parallel with one another [2]. A comprehensive examination of various types of swarms is presented in [3].
The field of System Identification (SID) has generated significant academic interest, as it encompasses the study of systems to unveil and model the interactions and dynamics in animal or robot swarms. A deeper understanding of these dynamics can aid in the advancement of more advanced swarm systems.\\
Identification algorithms have been proposed that integrate structure determination with parameter estimation using an orthogonal least squares approach [4]. Determining the parameters through estimation of causal entropy can provide insight into the impact of components in multivariate time series data [5]. Another non-parametric method, entropic regression, has been utilized to estimate the parameters of dynamic equations [6].
Recently, various learning-based methods have also been developed for SID. Temporal Convolutional Neural Networks (CNNs) utilize masked convolutions to implement a fixed-range sequence model [7]. Additionally, multi-step neural network approaches, such as Adams-Bashforth [8], Cluster-Networks [9], and Physics-Informed Deep Learning models [10] and [11], have been proposed. Another study evaluates the use of machine learning models to supplement knowledge-based mathematical models in cases where analytical models may fail, for example, due to unknown noise models or approximations [12].\\
A recent development in the field of nonlinear system identification has introduced a new method for uncovering the structure of a nonlinear dynamic system from data. This approach utilizes symbolic regression to determine the system's dynamics and strikes a balance between the model's complexity (number of terms) and its agreement with the data. However, this regression problem can be computationally expensive and may not scale well to large-scale systems, and may result in over-fitting. Other techniques for discovering emergent behavior and governing equations from time-series data include automated statistical inference of dynamics, equation-free modeling, and empirical dynamic modeling. One method, Sparse Identification of Nonlinear Dynamics (SINDy), produces a sparse, nonlinear regression that automatically identifies the relevant terms in the system from a library of functions [13].

\section{Methods and Approaches for System Identification}

Simulations were performed to generate data, utilizing equation (15) in Python. The resulting data comprised of the position and velocity of each swarm particle for N timesteps. A representative example of the generated data, presented in CSV format, can be found in the appendix.
The system's steady state is characterized by the time derivatives of its parameters being equal to and maintaining zero values. The eventual attainment of the steady state may vary depending on the system's initial conditions and the path or duration required to reach it. To comprehend the system more thoroughly, the methodologies applied should be categorized into four distinct training approaches.
\\
\\
\\
\begin{enumerate}
\item
    Training on Transient Dynamics with Identical Initial Conditions to Test Data
 \item   
    Training on Transient Dynamics with Dissimilar Initial Conditions to Test Data
\item
    Training on Steady State Dynamics with Identical Initial Conditions to Test Data
\item
    Training on Steady State Dynamics with Dissimilar Initial Conditions to Test Data
\end{enumerate}
The baseline techniques were executed and evaluated as described above. The models were trained through a folding approach, where a consecutive set of observations were utilized to predict the subsequent step. For instance, the first five steps were utilized to forecast the sixth step, the second to sixth steps predicted the seventh step, and so forth. This resulted in a single batch shape, assuming a swarm of 32 particles, as follows:
\begin{align}
    5\times 32\times 4\\
    5\times 128
\end{align}
In order to assess and compare our models, we defined the Mean Field Error (MFE) as a metric. The MFE quantifies the discrepancy between the mean positions of the predicted swarm particles and the ground truth particles. It is computed as the Euclidean distance between the mean positions of the swarm particles, providing insight into the stability achieved by the model and its ability to learn from the time-series data. The calculation of the MFE is described as follows:
\begin{align}
    MFE = \sqrt{(x_{m,true}-x_{m,pred})^2+(y_{m,true}-_{m,pred})^2}
\end{align}
\section{Non Deep learning Models}

This section outlines one method: a Regression Model-based Forecasting approach for predicting swarm behavior for deriving the equation form of the swarm's dynamical system.

\subsection{Least Square-Regression Model}

The data utilized to construct models predicting the future behavior of swarms is time-series data. For basic time-series forecasting, Linear Regression Models are commonly used. Despite the swarm system being non-linear, linear regression models can locally approximate the non-linear function and still be employed for forecasting future samples, provided the local neighborhood of the estimation is small. One of the initial models established is a Linear Regression Model based on Least Squares. The Ordinary Least Squares (OLS) method is employed to evaluate the various regression lines, where the regression line that minimizes the sum of the squared differences between the observed and predicted dependent variables is selected according to OLS.

Given a set of m samples, $x = [x_{1},x_{2},x_{3},...,x_{m}]$, the model will learn a weight vector, w, to predict future samples, $\tilde{x} = [x_{m+1},x_{m+2},x_{m+3},...,x_{m+n}]$, as $\tilde{x} = w^{T}x$, where the predicted samples are n steps ahead of the current time. The quantity m represents the number of in-samples, while n represents the number of out-samples that the model predicts. The optimal feature set size (m) and optimal prediction horizon (n) are generally determined through cross-validation and serve as hyperparameters for the model.
\begin{enumerate}
\item
Model for Steady State Dynamics: A swarm model was generated using the defined model from Section 2, producing data over 3000 time steps. A test-train split was created, using 2000 samples for training and 1000 samples for testing. The data primarily represents the steady state behavior of the swarm system. A regression model was built, taking 10 samples as input and predicting the 11th sample. After training, a sliding window approach was used for testing the model. The model was initially given the last 10 samples of the training data to predict a new sample, then the window was shifted to include the predicted sample to predict additional samples until 1000 samples were predicted.
\item
A Regression Model for Transient Dynamics: The swarm model data consisting of 250 time steps was generated as per the procedure outlined in Section 2. A test-train split was established, with 150 samples designated for training and 100 samples reserved for testing, to primarily capture the transient dynamics of the swarm system. A regression model was built, which employed 10 samples to predict the 11th sample. After training, a sliding window approach was employed for evaluating the model's performance. The last 10 samples of the training data were initially fed to the model, which predicted the next sample. This process was repeated by folding ahead and incorporating the predicted sample until 100 samples were predicted.
\end{enumerate}

\begin{figure}[H]
\centering
\includegraphics[width=3.34in]{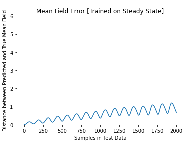}
\caption{MFE for Steady State Predictions}
\end{figure}
The study finds that a regression model with an optimal feature size and prediction horizon of 10 and 1 respectively can accurately predict the steady state evolution of a swarm system with minimal mean field error (MFE). As the number of prediction samples increases, the MFE begins to rise. The model can predict the steady state behavior of the swarm up to approximately 1000 steps ahead with acceptable MFE. However, its performance on transient data is poor due to its inability to capture the highly non-linear nature of the swarm system. This linear regression model does not provide insight into the underlying dynamics of the swarm system as it is linear in nature and the swarm system is highly non-linear. The model's success in predicting steady state may be due to the linear stability of the multi-agent system in the steady state.
\begin{figure}[H]
\centering
\includegraphics[width=3in]{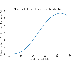}
\caption{MFE for Transient Predictions}
\end{figure}

\begin{figure}[]
\centering
\includegraphics[width=3in]{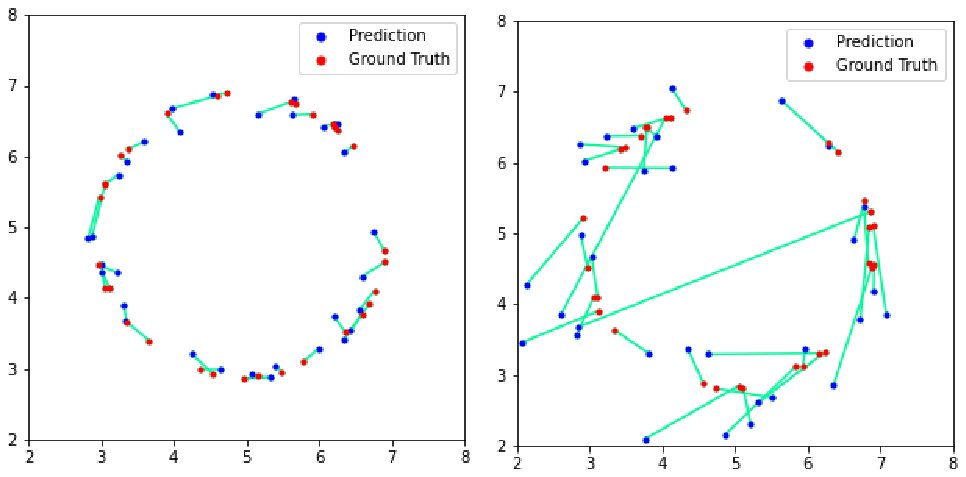}
\caption{Prediction for Steady State}
\end{figure}
\begin{figure}[]
\centering
\includegraphics[width=3in]{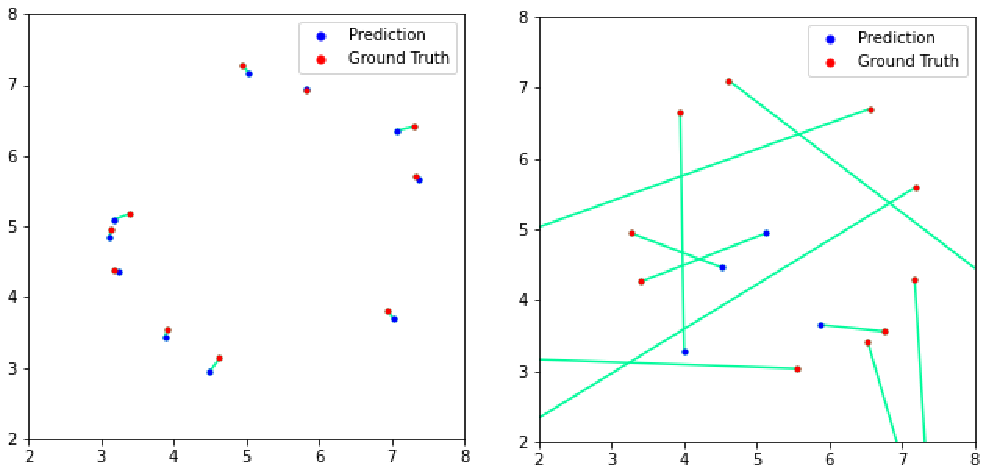}
\caption{Prediction for Transient}
\end{figure}

\section{Deep Learning Models}

The study evaluated three deep learning baselines, including a standard feed forward network, recurrent neural networks (RNNs), and temporal convolutional neural networks (CNNs). The RNNs and temporal CNNs demonstrated superior performance in extrapolating steady swarm dynamics and accurately matching the ground truth. However, when trained on transient data, these models failed to extrapolate and reach stability, revealing their limitations in capturing the inherent dynamics of the system. This highlights the strengths of these models in inferring temporal dependencies but not in learning the underlying dynamics of the system.

\subsection{Multi-Layer Perceptron }

The state dynamics of the system is characterized by a set of non-linear equations. To capture this non-linearity, a baseline was implemented using a fully connected neural network. This network, with appropriate activation functions between layers, has the potential to capture the temporally dependent dynamics of the swarm. The input to the network is a tensor consisting of positional and velocity data of 32 particles over five time steps, and the objective of the model is to predict the state and velocity of the swarm for the next time step. The following table presents the final hyperparameters that produced the most reliable outcomes:

\begin{table}[H]
\begin{center}
\begin{tabular}{l|l}
\hline
Hidden Layer Size           & 256   \\ \hline
Optimizer                   & SGD   \\ \hline
Learning Rate               & 0.001 \\ \hline
Training Window for Dataset & 5     \\ \hline
\end{tabular}
\caption{Hyperparameters for MLP}
\label{tab:my-table1}
\end{center}
\end{table}

\subsection{Recurrent Neural Network}
Recurrent Neural Networks (RNNs) are extensions of feedforward neural networks with a built-in memory component. They are recurrent in nature as they perform the same computation for every sequence of input data, with the output dependent on the previous computation. The output is then copied and fed back into the network, allowing it to consider both the current input and the previous output in making a decision. This makes RNNs highly effective in identifying patterns in time-series data or data where the samples at a particular time are assumed to be dependent on the preceding sample.
\begin{figure}[H]
\centering
\includegraphics[width=3.34in]{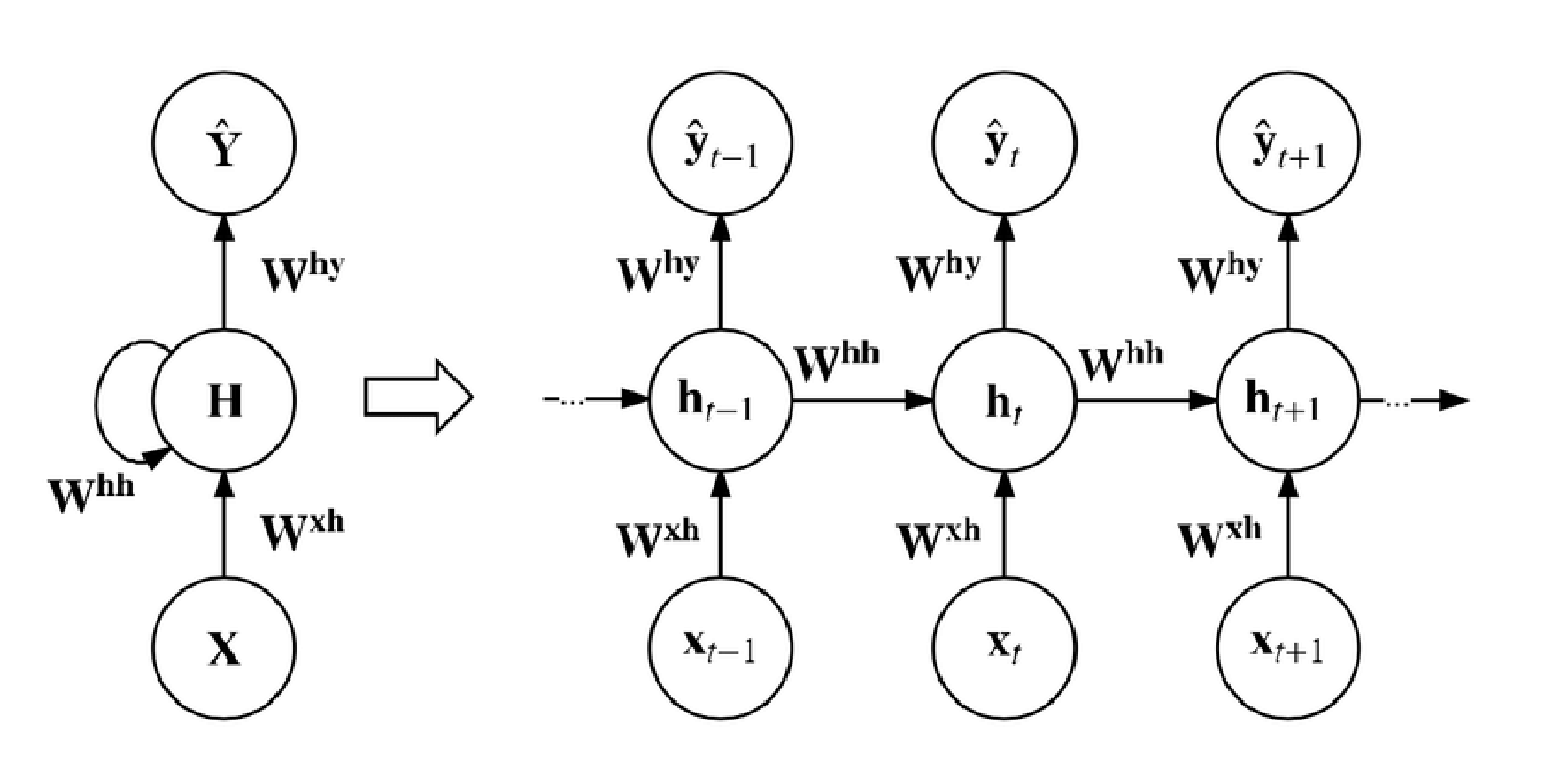}
\caption{RNN Unrolled Structure}
\end{figure}
A single-layer RNN was employed, followed by a linear activation function. The loss function used was nn.MSELoss(). The hyperparameters optimized included the hidden size, the number of RNN layers, the optimizer, and the learning rate, as detailed below:
\begin{table}[H]
\begin{center}
\begin{tabular}{l|l}
\hline
Hidden Layer Size           & 256   \\ \hline
Number of RNN layers           & 1   \\ \hline
Optimizer                   & SGD   \\ \hline
Learning Rate               & 0.005 \\ \hline
Training Window for Dataset & 5     \\ \hline
\end{tabular}
\caption{Hyperparameters for RNN}
\label{tab:my-table2}
\end{center}
\end{table}
\subsection{Convolutional Neural Networks}

While Recurrent Neural Networks (RNNs) and their derivatives such as Long Short-Term Memories (LSTMs) and Gated Recurrent Units (GRUs) have traditionally been the preferred models for time-series based data, Convolutional Neural Networks (CNNs) have also been effectively applied to the learning of time-series data as described in [14]. Given the presence of multiple swarm particles with underlying interaction forces in the data, incorporating a form of convolution may be beneficial in capturing both local and temporal information. In this implementation, 1-dimensional convolutions were used instead of 2-dimensional convolutions. The following presents the final architecture of our implemented model:
\begin{figure}[H]
\centering
\includegraphics[width=3.34in]{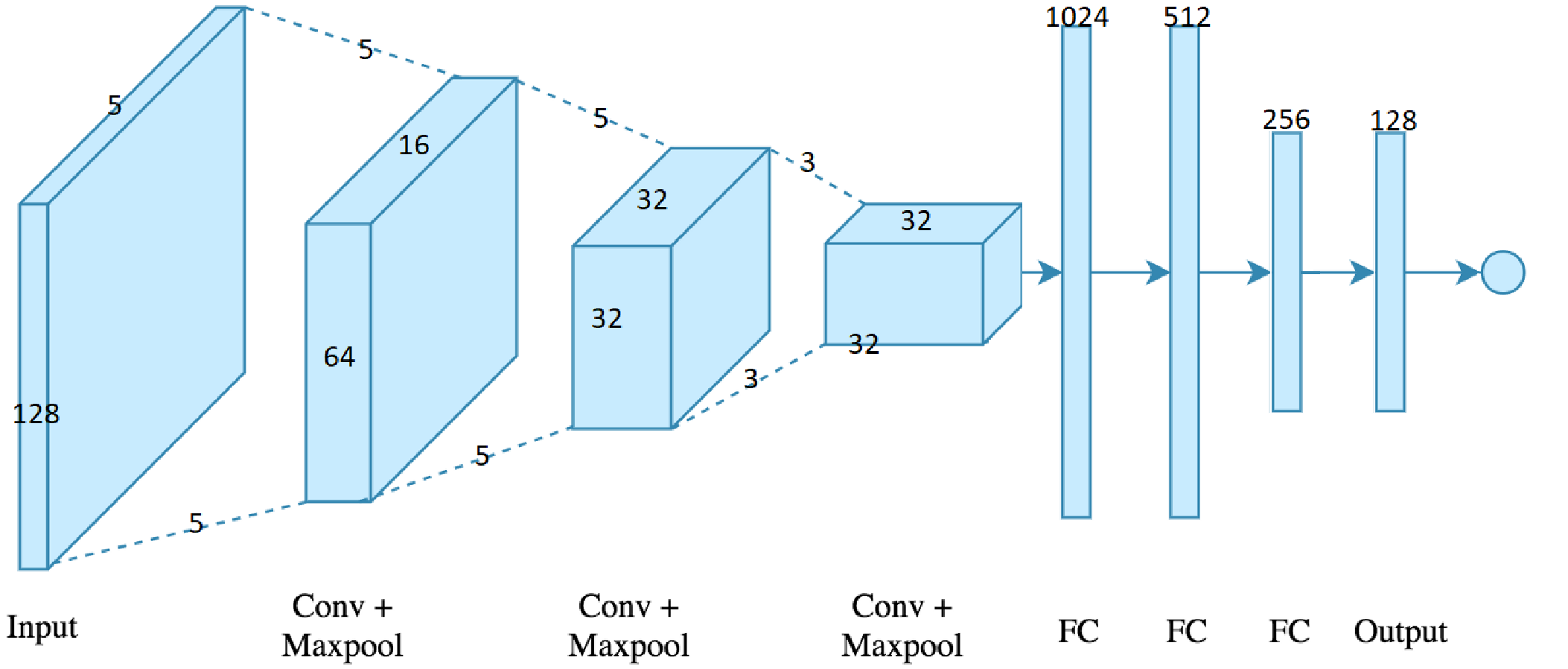}
\caption{Implemented CNN Architecture}
\end{figure}

Several architectural and filtering configurations were evaluated, with the final configuration producing the most consistent results. The use of Max-Pooling, Adaptive Average-Pooling, and Dropout resulted in suboptimal performance, hence not included in the final architecture. The ReLU activation function was applied to most layers. The chosen loss function was Mean Square Error $(nn.MSE)$. Both Stochastic Gradient Descent $(SGD)$ and Adam optimization algorithms were experimented with. Although Adam demonstrated fast convergence at times, its performance was inconsistent across different learning rates. On the other hand, SGD consistently converged quickly at a learning rate of $5 \times 10^{-3}$ and was chosen as the final optimizer and learning rate.
\begin{table}[H]
\begin{center}
\begin{tabular}{l|l}
\hline
Loss Function           & Mean Square Error   \\ \hline
Optimizer                   & SGD   \\ \hline
Learning Rate               & 0.0005 \\ \hline
Training Window for Dataset & 5     \\ \hline
\end{tabular}
\caption{Hyperparamters for CNN}
\label{tab:my-table3}
\end{center}
\end{table}

\begin{table}[H]
\begin{center}
\begin{tabular}{|ll|ll|ll|}
\multicolumn{2}{|l|}{First Lyer}    & \multicolumn{2}{l|}{Second Layer}  & \multicolumn{2}{l|}{Third Layer}   \\ \hline
\multicolumn{1}{|l|}{Filter 1}  & 5 & \multicolumn{1}{l|}{Filter 2}  & 5 & \multicolumn{1}{l|}{Filter 3}  & 3 \\
\multicolumn{1}{|l|}{Padding 1} & 2 & \multicolumn{1}{l|}{Padding 2} & 2 & \multicolumn{1}{l|}{Padding 3} & 1 \\
\multicolumn{1}{|l|}{Stride 1}  & 2 & \multicolumn{1}{l|}{Stride 2}  & 2 & \multicolumn{1}{l|}{Stride 3}  & 1
\end{tabular}
\caption{Parameters of Convolution Layers}
\label{tab:my-table4}
\end{center}
\end{table}

The performance of the models was inadequate when extrapolating and predicting transient swarm dynamics. These findings suggest the need for further research in developing robust deep learning solutions capable of generalizing the system's dynamics and accurately forecasting the future trajectories of swarm particles during transient states.

\subsection{Swarm Dynamics Predictive Performance through Neural Network Models}

The Multi-Layer Perceptron (MLP) model was initially evaluated as a baseline. The MLP was capable of capturing the non-linear properties of the model. However, its performance was highly contingent on the initial conditions of the particles, resulting in a marked decrease in its ability to predict the true test trajectories of the swarm when these conditions deviated from the training data. The comparison between the performance of the model on steady-state data and data consisting of both transient and steady-state dynamics revealed a substantial reduction in performance during the latter case, particularly during the transient phase, as indicated by the Mean Field Errors. These findings suggest that the MLP is not capable of uncovering the underlying behavior of the swarm and instead focuses solely on minimizing the Mean Field Error. As a result, it is prone to failure when initial conditions differ or during transient scenarios. In addition, the training time required by the MLP was significantly longer (500 epochs) compared to the RNN and CNN models (50 epochs), further hindering its utility. Given these limitations, the MLP was not tested as extensively as the CNN and RNN models.
\begin{figure}[H]
\centering
\includegraphics[width=3.34in]{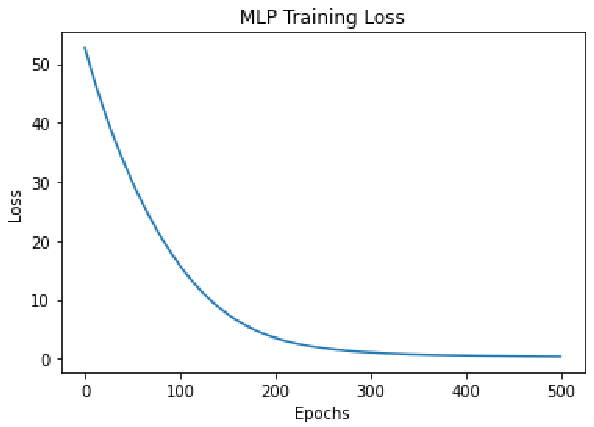}
\caption{Training loss for model trained on Steady State for MLP}
\end{figure}
The RNN and CNN models exhibited efficient performance during the training process, exhibiting rapid convergence across the four methodologies outlined in the Methods section. Experiments with various epoch lengths revealed consistent results for lengths exceeding 50 epochs. The loss plots for the two models in one case are depicted in Figures 8 and 9.
\begin{figure}[H]
\centering
\includegraphics[width=3.34in]{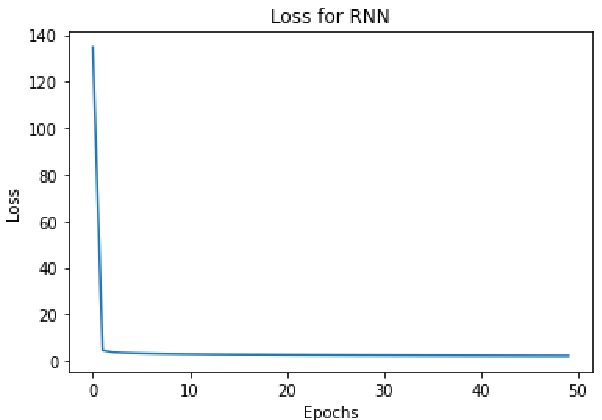}
\caption{Training loss for model trained on Steady State for RNN}
\end{figure}
\begin{figure}[H]
\centering
\includegraphics[width=3.34in]{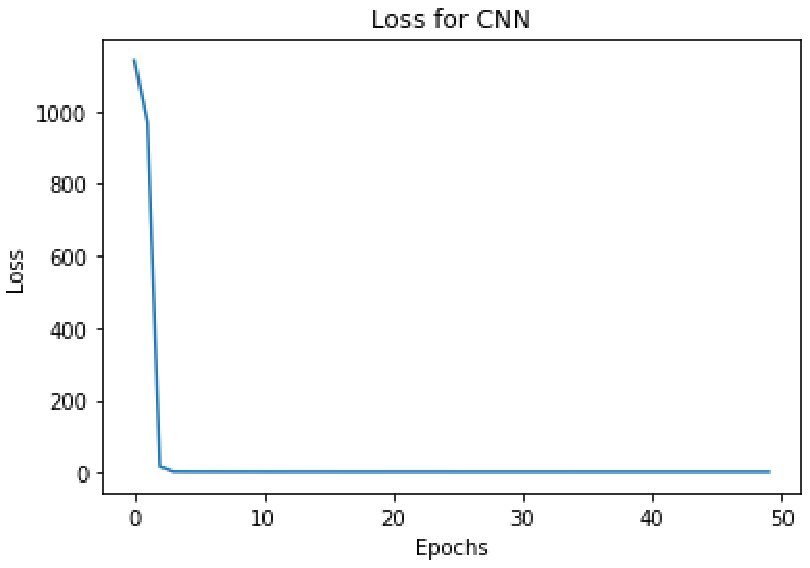}
\caption{Training loss for model trained on Steady State for CNN}
\end{figure}
Despite their rapid convergence to a minimal cost (within 50 epochs), the RNN and CNN models only demonstrated exceptional performance under one methodology, namely when they were trained on steady-state data with initial conditions identical to the test data. The results were inconsistent for the other training methodologies, as illustrated by the Mean Field Error plots.
As demonstrated in Figures 10 and 11, no clear better-performing model emerges, owing to the presence of a transient state in the evaluation data. As previously noted, the model trained on steady-state data was expected to perform well, and it did once the transient state was removed, causing a significant change in the Mean Field Error. This suggests that both the RNN and CNN models can only learn the stable part of the swarm system. The plots for all other truncated tests are included in the appendix. Although RNN and CNN seem to have comparable performance, a closer examination of the predicted data from each model via animation reveals that while the Mean Field Error is similar for both models, the CNN model exhibits a noticeable phase shift.
\begin{figure}[H]
\centering
\includegraphics[width=3.34in]{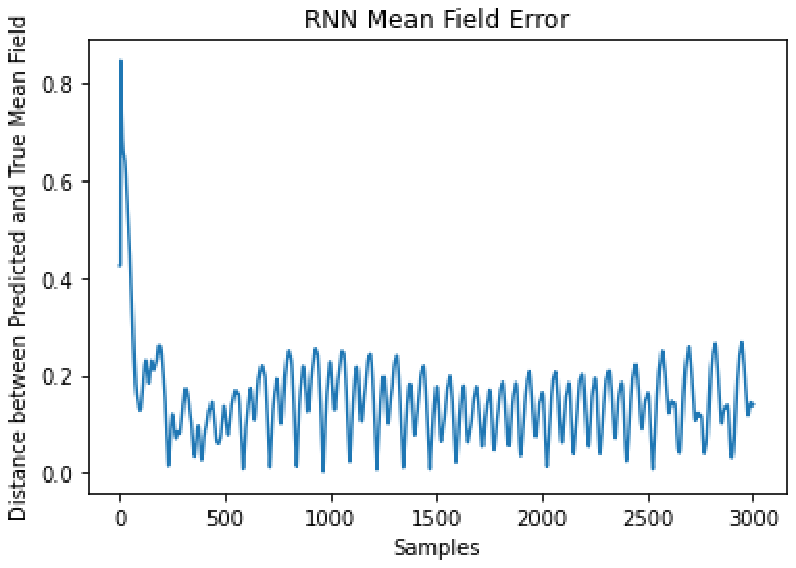}
\caption{trained on Steady State for RNN MFE}
\end{figure}
\begin{figure}[H]
\centering
\includegraphics[width=3.34in]{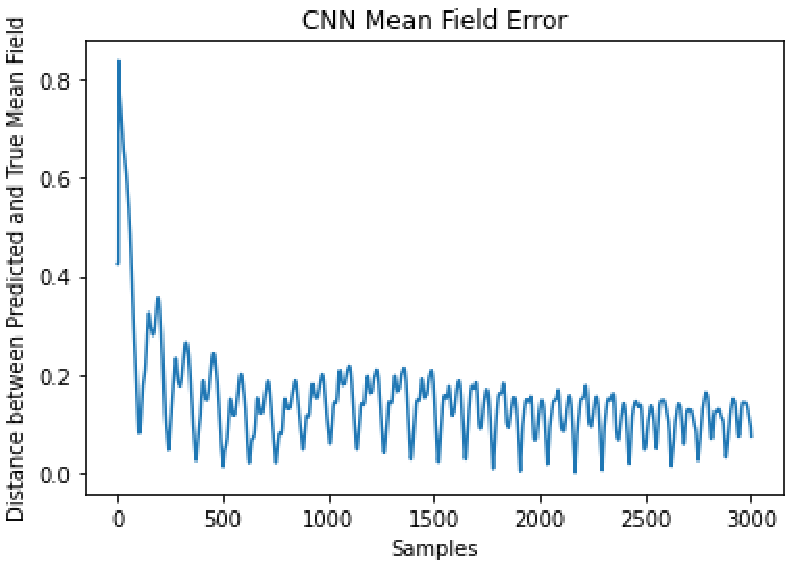}
\caption{trained on Steady State for CNN MFE}
\end{figure}
The analysis of the results presented in this section indicates that the RNN and MLP models have an advantage in learning inter-particle interactions and individual particle behaviors compared to only learning the steady state of the entire system, which is the case with the CNN model. The three deep-learning-based models (MLP, RNN, and CNN) are capable of learning steady-state models from data with the same initial conditions, but they are unable to generalize well when dealing with the presence of transient states or states with different initial conditions. The results suggest that the deep-learning-based models, similar to the non-baseline models, are not able to identify the hidden state dynamics of the swarm system and struggle with capturing the highly non-linear properties of the transient system.
\section{Advanced Deep Learning Model - Neural ODE}
The proposed model boasts two distinctive features that enhance its efficacy. Firstly, it adopts the concept of Neural ODE, a non-architectural but innovative method for optimizing through the utilization of an ODE solver. Backpropagation through an ODE solver is feasible, but it necessitates the storage of variables at each time step, leading to substantial memory overhead. This model circumvents this issue by defining adjoint states, the time dynamics of which can be obtained through numerical integration with the ODE solver. These adjoint states then facilitate the calculation of derivatives of each parameter with respect to the loss function, thereby enabling the ODE solver to be utilized as a "black box", regardless of the specific ODE solver employed.

Our objective is to approximate the function $\dot{X}=f(X)$ as a neural network $\hat{f}$ with parameters $\theta$ , using the form presented in equation 15. The optimization target is to minimize the scalar-valued loss function L, given by $L(ODESolve(X(t_{0}),\hat{f},t_{0},t_{1},\theta))$. The definition of adjoint states and the calculation of their time dynamics are detailed in [15].
In this model, the selection of the neural network architecture utilized for approximating the function $\hat{f}$ is the second aspect considered. Given that the dynamics of $\dot{X}$ is the direct target for approximation, physical characteristics of the swarm system can be leveraged to inform the choice of the neural network. The homogeneity of the observed swarm implies that the dynamics of each agent must be uniform and that the interactions between agents must also exhibit consistency. In light of these considerations, the architecture was designed as depicted in Figure 12. The utilization of a physics-informed network significantly shrinks the parameter space, thereby expediting the training process.
\begin{table}[H]
\begin{center}
\begin{tabular}{l|l|l|l|}
                                            & Intrinsic & Interaction & Aggregation \\ \hline
\multicolumn{1}{|l|}{Number of Layers}      & 3               & 3                 & 3                 \\
\multicolumn{1}{|l|}{Nonlinearity Input} & Cubic    & None              & None             
\end{tabular}
\caption{Physics-informed model architecture}
\label{tab:my-table5}
\end{center}
\end{table}

\begin{table}[H]
\begin{center}
\begin{tabular}{l|l}
\hline
ODESolver Type & Euler          \\ \hline
ODESolver Step & 0.05           \\ \hline
Optimizer      & Adam(lr=0.01) \\ \hline
Loss Function  & MSE            \\ \hline
\end{tabular}          
\caption{Physics-informed model hyperparameters}
\label{tab:my-table6}
\end{center}
\end{table}
Tables 5 and 6 present the architecture of the final model and the hyperparameters employed for the training process, respectively.
\begin{figure}[H]
\centering
\includegraphics[width=3.34in]{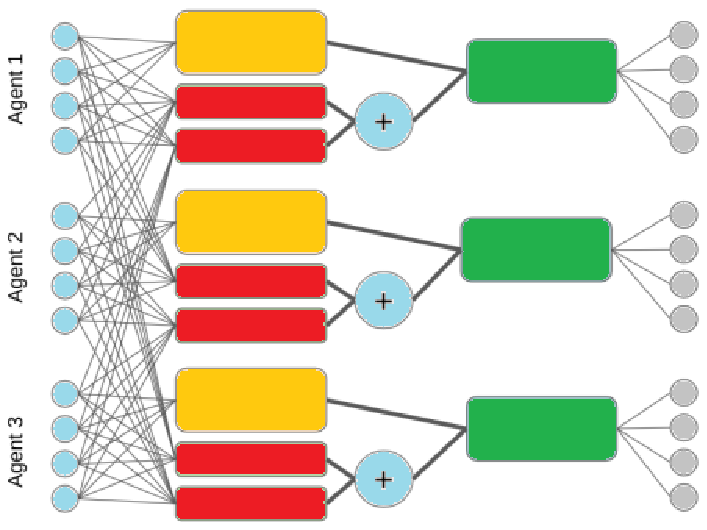}
\caption{Physics-informed model for 3 agent swarm team}
\end{figure}
\begin{figure}[H]
\centering
\includegraphics[width=3.34in]{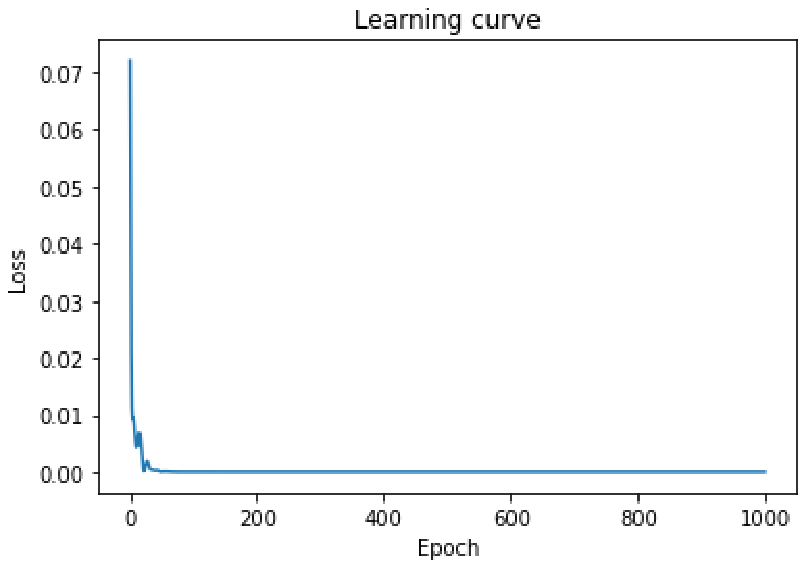}
\caption{Learning curve of NODE for 3 agent swarm team}
\end{figure}
\begin{figure}[H]
\centering
\includegraphics[width=3.34in]{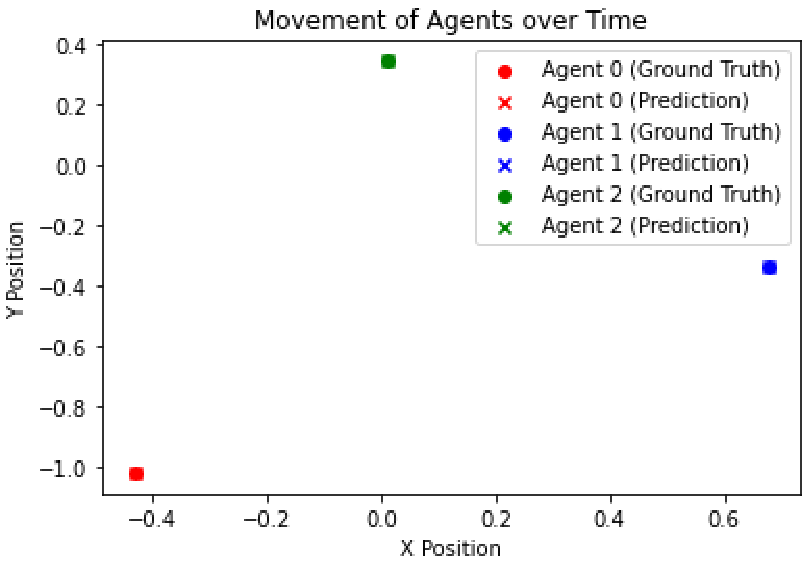}
\caption{Predicted trajectory of agents compare to ground truth}
\end{figure}
\section{Conclusions}
In this study, we investigated the effectiveness of various neural network models in capturing the dynamics of nonlinear Ordinary Differential Equations (ODEs) systems. Our aim was to evaluate the performance of existing deep learning and non-deep learning models in learning the stability of swarm systems through time-series data. The results showed that traditional deep learning models and non-deep learning baselines were not robust enough to handle the nonlinearity and transient states of the swarm systems. The models performed well when evaluating systems with the same initial conditions but struggled with systems with different initial conditions and transient states.

This led us to explore a new deep learning method, Neural ODE, which provides a framework for training ODEs within larger models. Despite finding a working model, it remains unclear how the network, in conjunction with Neural ODE, captures the true dynamical process of the system. This is because the presence of bifurcations and catastrophes in dynamical systems can cause instability and hysteresis, hindering the learning of a good model. Further research is needed to fully understand the limitations and capabilities of Neural ODEs in capturing the nonlinear dynamics of swarm systems.

\section*{References}

\begin{enumerate}
  \item C. W. Reynolds, "Flocks, herds and schools: A distributed behavioral model," in \emph{Proceedings of the 14th annual conference on Computer graphics and interactive techniques}, 1987, pp. 25-34.

  \item Csaba Virágh, Gábor Vásárhelyi, Norbert Tarcai, Tamás Szőrényi, Gergő Somorjai, Tamás Nepusz, and Tamás Vicsek, "Flocking algorithm for autonomous flying robots," \emph{Bioinspiration \& biomimetics}, vol. 9, no. 2, p. 025012, 2014.

  \item Carl Kolon and Ira B Schwartz, "The dynamics of interacting swarms," \emph{arXiv preprint arXiv:1803.08817}, 2018.

  \item Sheng Chen, Stephen A Billings, and Wan Luo, "Orthogonal least squares methods and their application to non-linear system identification," \emph{International Journal of control}, vol. 50, no. 5, pp. 1873-1896, 1989.

  \item Pileun Kim, Jonathan Rogers, Jie Sun, and Erik Bollt, "Causation entropy identifies sparsity structure for parameter estimation of dynamic systems," \emph{Journal of Computational and Nonlinear Dynamics}, vol. 12, no. 1, 2017.

  \item Abd AlRahman R AlMomani, Jie Sun, and Erik Bollt, "How entropic regression beats the outliers problem in nonlinear system identification," \emph{Chaos: An Interdisciplinary Journal of Nonlinear Science}, vol. 30, no. 1, p. 013107, 2020.

  \item John M Maroli, Ümit Özgüner, and Keith Redmill, "Nonlinear system identification using temporal convolutional networks: a silverbox study," \emph{IFAC-PapersOnLine}, vol. 52, no. 29, pp. 186-191, 2019.

  \item Maziar Raissi, Paris Perdikaris, and George Em Karniadakis, "Multistep neural networks for data-driven discovery of nonlinear dynamical systems," \emph{arXiv preprint arXiv:1801.01236}, 2018.

  \item Rong-Jong Wai and Alex S Prasetia, "Adaptive neural network control and optimal path planning of UAV surveillance system with energy consumption prediction," \emph{Ieee Access}, vol. 7, pp. 126137-126153, 2019.

  \item Maziar Raissi, Paris Perdikaris, and George Em Karniadakis, "Physics informed deep learning (part i): Data-driven solutions of nonlinear partial differential equations," \emph{arXiv preprint arXiv:1711.10561}, 2017.

  \item Tayyab Manzoor, Hailong Pei, Zhongqi Sun, and Zihuan Cheng, "Model Predictive Control Technique for Ducted Fan Aerial Vehicles Using Physics-Informed Machine Learning," \emph{Drones}, vol. 7, no. 1, p. 4, 2022.

  \item Jaideep Pathak, Alexander Wikner, Rebeckah Fussell, Sarthak Chandra, Brian R Hunt, Michelle Girvan, and Edward Ott, "Hybrid forecasting of chaotic processes: Using machine learning in conjunction with a knowledge-based model," \emph{Chaos: An Interdisciplinary Journal of Nonlinear Science}, vol. 28, no. 4, p. 041101, 2018.

  \item Kathleen Champion, Bethany Lusch, J Nathan Kutz, and Steven L Brunton, "Data-driven discovery of coordinates and governing equations," \emph{Proceedings of the National Academy of Sciences}, vol. 116, no. 45, pp. 22445-22451, 2019.

  \item Hassan Ismail Fawaz, Germain Forestier, Jonathan Weber, Lhassane Idoumghar, and Pierre-Alain Muller, "Deep learning for time series classification: a review," \emph{Data mining and knowledge discovery}, vol. 33, no. 4, pp. 917-963, 2019.

  \item Han Zhang, Xi Gao, Jacob Unterman, and Tom Arodz, "Approximation capabilities of neural ordinary differential equations," \emph{arXiv preprint arXiv:1907.12998}, vol. 2, no. 4, pp. 3-1, 2019.
\end{enumerate}




% ...


\bibliography{asme2e}



\end{document}